\documentclass[journal]{IEEEtran}
\usepackage{blindtext,booktabs,amsmath}
\usepackage[]{color}
\usepackage{amsmath}
\usepackage{multirow}
\usepackage{hhline}
\usepackage[]{caption}
\usepackage[caption=false]{subfig}
\usepackage[english]{babel}
\usepackage[utf8]{inputenc}
\usepackage{algorithm}
\usepackage{algorithmic}
\usepackage{multirow,tabularx}

\usepackage{amssymb}

\usepackage{array}
\usepackage{url}

\newcommand{\argmin}{\arg\!\min}

\newcommand{\abs}[1]{\left|#1\right|}

\ifCLASSINFOpdf
  \usepackage[pdftex]{graphicx}
  \graphicspath{{./pdf/}{./jpeg/}{./png/}}
  \DeclareGraphicsExtensions{.pdf,.jpeg,.png}
\else
  \usepackage[dvips]{graphicx}
  \graphicspath{{./eps/}}
  \DeclareGraphicsExtensions{.eps}
\fi

\begin{document}
\title{Extraction of V2V Encountering Scenarios from Naturalistic Driving Database
}


\author{Zhaobin Mo, Sisi Li, Diange Yang, Ding Zhao

\thanks{* The first two authors, Z. Mo and S. Li, have made equal contributions to this work.}
\thanks{Z. Mo is with the Automotive Engineering at the Tsinghua University, Beijing, China, 100084.}%
 \thanks{S. Li is with the Robotics Institute, University of Michigan, Ann Arbor, MI, 48109}
 \thanks{D. Yang is with the Automotive Engineering at the Tsinghua University, Beijing, China, 100084.}%
 \thanks{D. Zhao is with the Department of Mechanical Engineering, University of Michigan, Ann Arbor,
 Ann Arbor, MI, 48109. (corresponding author: Ding Zhao zhaoding@umich.edu)}
 \thanks{This work was supported by National Natural Science Foundation of China under grant No. 61773234 and the International Sience \& Technology Cooperation Program of China under contract No. 2016YFE0102200}

}


\maketitle

\begin{abstract}
It is necessary to thoroughly evaluate the effectiveness and safety of Connected Vehicles (CVs) algorithm before their release and deployment. Current evaluation approach mainly relies on simulation platform with the single-vehicle driving model. The main drawback of it is the lack of network realism. To overcome this problem, we extract naturalistic V2V encounters data from the database, and then separate the primary vehicle encounter category by clustering. A fast mining algorithm is proposed that can be applied to parallel query for further process acceleration. 4,500 encounters are mined from a 275 GB database collected in the Safety Pilot Model Program in Ann Arbor Michigan, USA. K-means and Dynamic Time Warping (DTW) are used in clustering. Results show this method can quickly mine and cluster primary driving scenarios from a large database. Our results separate the car-following, intersection and by-passing, which are the primary category of the vehicle encounter. We anticipate the work in the essay can become a general method to effectively extract vehicle encounters from any existing database that contains vehicular GPS information. What's more, the naturalistic data of different vehicle encounters can be applied in Connected Vehicles evaluation.
\end{abstract}

\begin{IEEEkeywords}
Connected Vehicle, clustering, data mining
\end{IEEEkeywords}

\IEEEpeerreviewmaketitle

\section{Introduction}

The connected vehicles (CVs) initiative enabled the introduction of mobile wireless communications technology into transportation safety, operations, and management \cite{agbolosuCVimpact}. Before the related algorithm's release and deployment, it must be evaluated thoroughly to ensure effectiveness and safety. 
There are two main fundamental questions regarding CV evaluation:

\begin{enumerate}
\item How to choose the driving model
\item How to choose the network configuration
\end{enumerate}

Existing research on CVs evaluation can be divided into two main categories. The first focuses on modeling and simulation of the driving model, and there has already been some mature simulation platform. \cite{agbolosuCVimpact}\cite{li2013estimating} used VISSIM as the simulation platform. The VISSIM deals with the stochastic nature of traffic by incorporating a range of parameters that use stochastic distribution, and pre-include traffic models like Wiedemann model. \cite{kari2014eco} used Simulation of Urban Mobility (SUMO) to control the simulation, it deals with the same problem also by pre-including some traffic model like intelligent driver model (IDM), Kerner’s three-phase model, and Wiedemann model. However, though these methods have been proven to simulate the vehicle driving precisely enough, it lacks realism of networks. This primary deficiency limits their application to a single site or local situation, like a particular intersection. 

The second category of research uses naturalistic data directly. \cite{2016microscopic} used the vehicle trajectories form the Next-Generation Simulation (NGSIM) project, a field-collected data set of vehicle movements along several corridors in the United States. The use of naturalistic traffic scenario data ensures the network realism. However, as it is collected from fixed cameras, the biggest drawback is the limited scope of traffic scenarios. Because of the difficulty of tracking multiple vehicles among different cameras, this method can only provide local information of a subset of the trajectories. What's more, the image information collected from cameras is vulnerable to the weather \cite{Sun2006On}.

\begin{figure}[t]
\includegraphics[width=\linewidth]{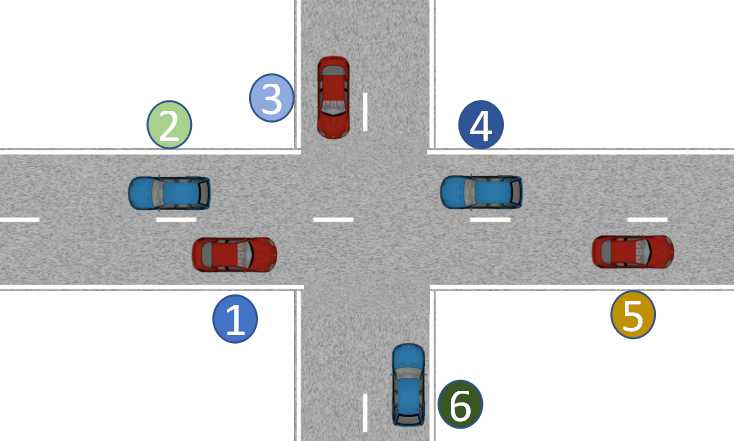}
     \caption{A multi-vehicle driving scenario with unsorted Vehicle Encounters. Intersection for vehicle combination 1-3, 1-6, 2-3, 2-6, 3-4, 3-5, 4-6, 5-6. Car-following for vehicle combination 1-5, 2-4. By-passing for vehicle combination 1-2, 1-4, 2-5, 3-6. Our job is to extract these binary combinations from  `chaotic' database.}
     \label{fig:intro_section}
 \end{figure}

A solution to the problems above is to use the on-board Global Position System (GPS). As an on-board sensor, GPS can track the position of the vehicle the whole trip, recording the complete trajectory for further analysis. Besides, this technique does not need expensive equipment and does not rely on vision or radar, hence performs well in most weather and light condition  \cite{knoop2017lane}. However, although GPS information is easy to get and there have already been some GPS databases \cite{haklay2008openstreetmap}, best to our knowledge, there are no data set of the scenario where vehicles are within the range of Dedicated Short Range Communications (DSRC). 

Most CV algorithms focus on specific driving scenarios before extending its application to all kinds of scenarios \cite{huang2017accelerated}. \cite{zhao2017lanechange} and \cite{zhao2017cf} used lane-change and car-following scenario for evaluation respectively. A scenario may contain too many vehicles and is complicated to analyze. Fig. \ref{fig:intro_section} is a typical driving scenario at a busy cross. In this scenario, there is a lot of combination of vehicles, thus generating many vehicle encounters. To better exploit and understand the driving scenario, it's necessary to find what primary encounter categories it contains, what is the proportion of different categories, and the inter-category envolving process. Our work focuses on the first step, finding the primary encounter categories from the massive and chaotic database. Our previous work used subjective rules to extract the scenarios \cite{zhao2017trafficnet} \cite{wang2018extracting}. However, these rules can not be applied to all scenarios. In this paper, we are using unsupervised learning method to extract the scenarios.

\begin{figure}[ht]
\includegraphics[width=\linewidth]{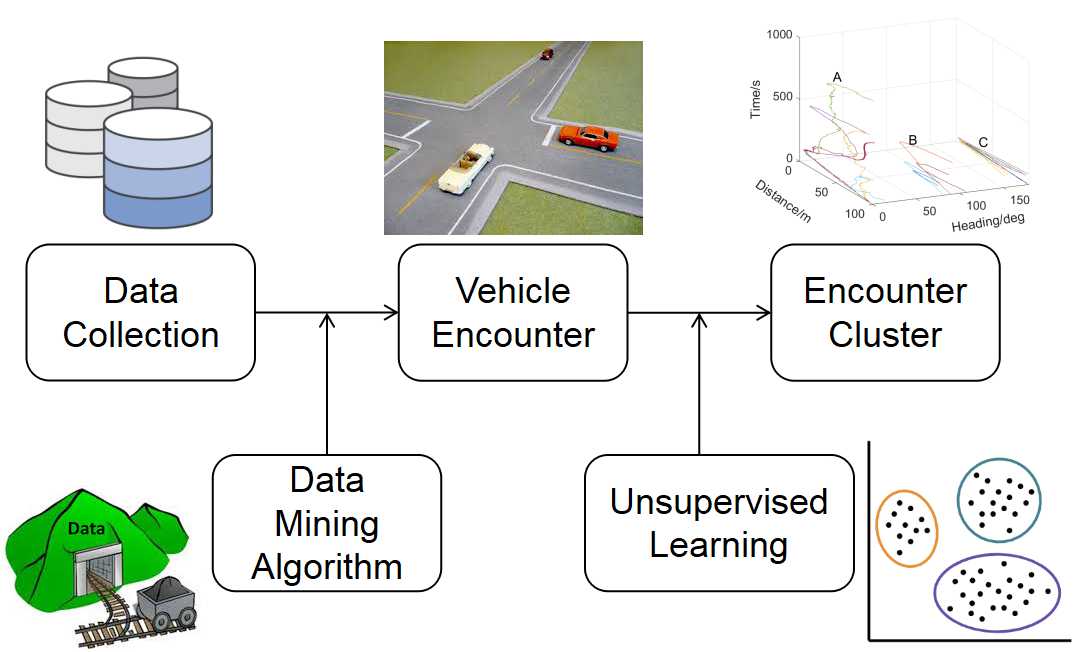}
     \caption{Procedure of research}
     \label{fig:procedure}
 \end{figure}

\begin{figure}[h]
\centering
\includegraphics[width=0.8\linewidth]{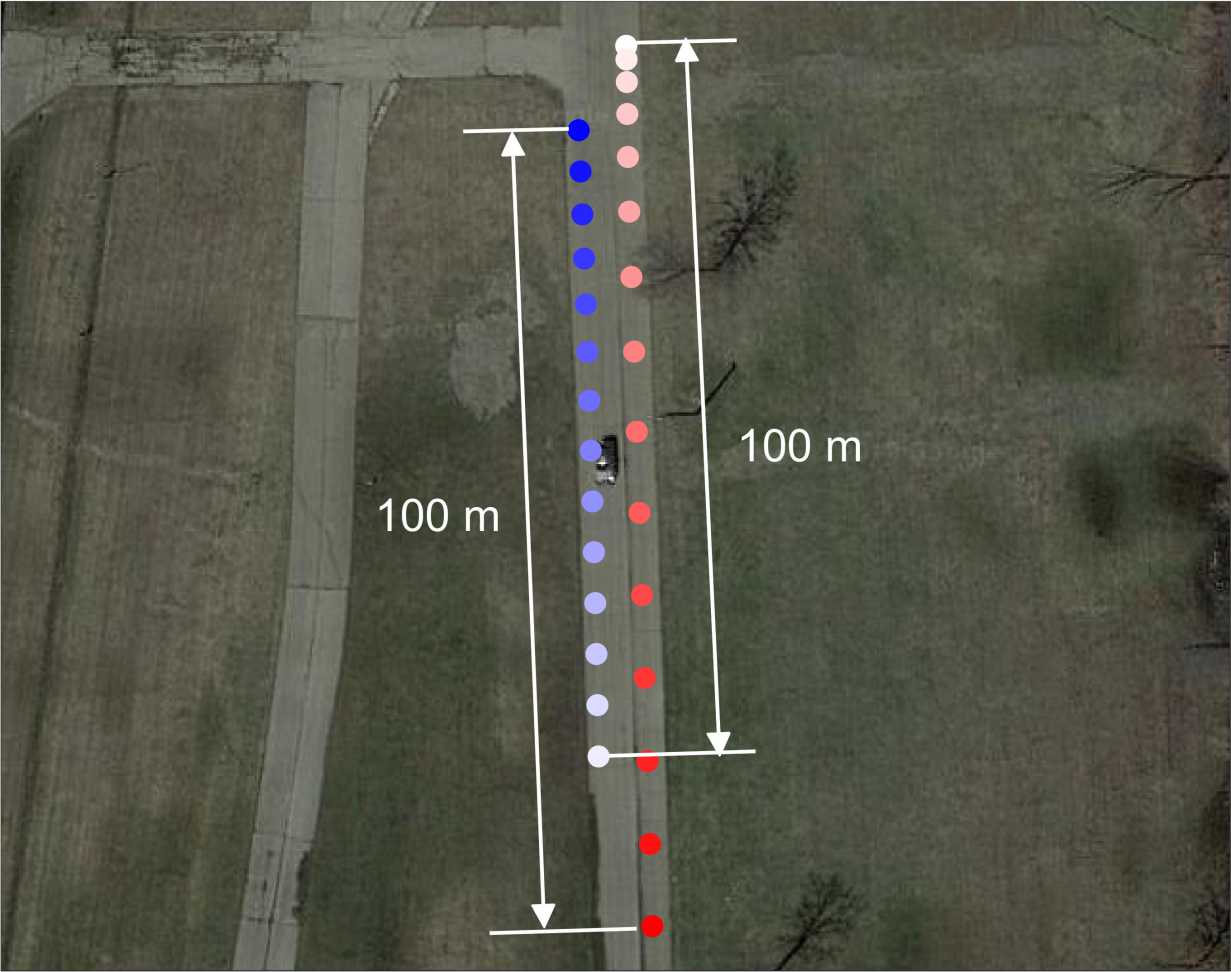}
     \caption{An example of Vehicle Encounters. Dark dot is start point and light dot is end point.}
     \label{fig:encounter}
 \end{figure}

In this work, we are using the SPMD (Safety Pilot: Model Deployment) database, which was conducted by the University of Michigan Transportation Research Institute (UMTRI) and provides 34 million miles driving data logged in the last four years in Ann Arbor area \cite{huang2017empirical}. The deployment includes approximately 2,800 equipped vehicles and 5 million trips in total. Latitude and altitude information is collected to track the position of each vehicle. Those data can be further processed for other parameters like velocity and heading angle. We start the research on the small part of the data ($\sim$ 295 GB). It comprises of 137 vehicles running for 3 years, and provides adequate information for this research. The data was collected with a sampling frequency of 10 Hz.
The vehicle encounter was defined as the scene where the vehicle distance is small than 100 m, as shown in Fig. \ref{fig:encounter}. The dots indicate the position of the vehicle at every sample time. 

The source code in this work can be accessed at https://github.com/zhao-lab/mo-extract-encounters-avec18.

\section{Mining Algorithm}
It is expected that we can extract the vehicle encounter as quickly as possible. To achieve that goal, one must handle the large scale of the database.
The data mining challenges concentrate on algorithm designs in tackling the difficulties raised by the Big Data volumes\cite{wu2014data}. This question is aggravated here as we are querying the combination of different vehicles, which shares a quadratic relationship with the vehicle number. This combinatorial explosion makes it unrealistic to query in a large database.

Another problem is the query criterion. Though the database is big, the vehicle encounter is scarce, and it is a waste of time and resource to calculate the vehicle distance for every possible combination at every time step. Thus A coarse filter is needed.

In this paper, we first preprocess and eliminate the abnormal data, then reorder the data to apply a queue structure for the inter-vehicle match. After that, we apply coarse filter methods to extract the potential encounters at the minimal cost. Last we finely filter the results to get the encounters.

\begin{figure}[ht]
\centering
\includegraphics[width=0.8\linewidth]{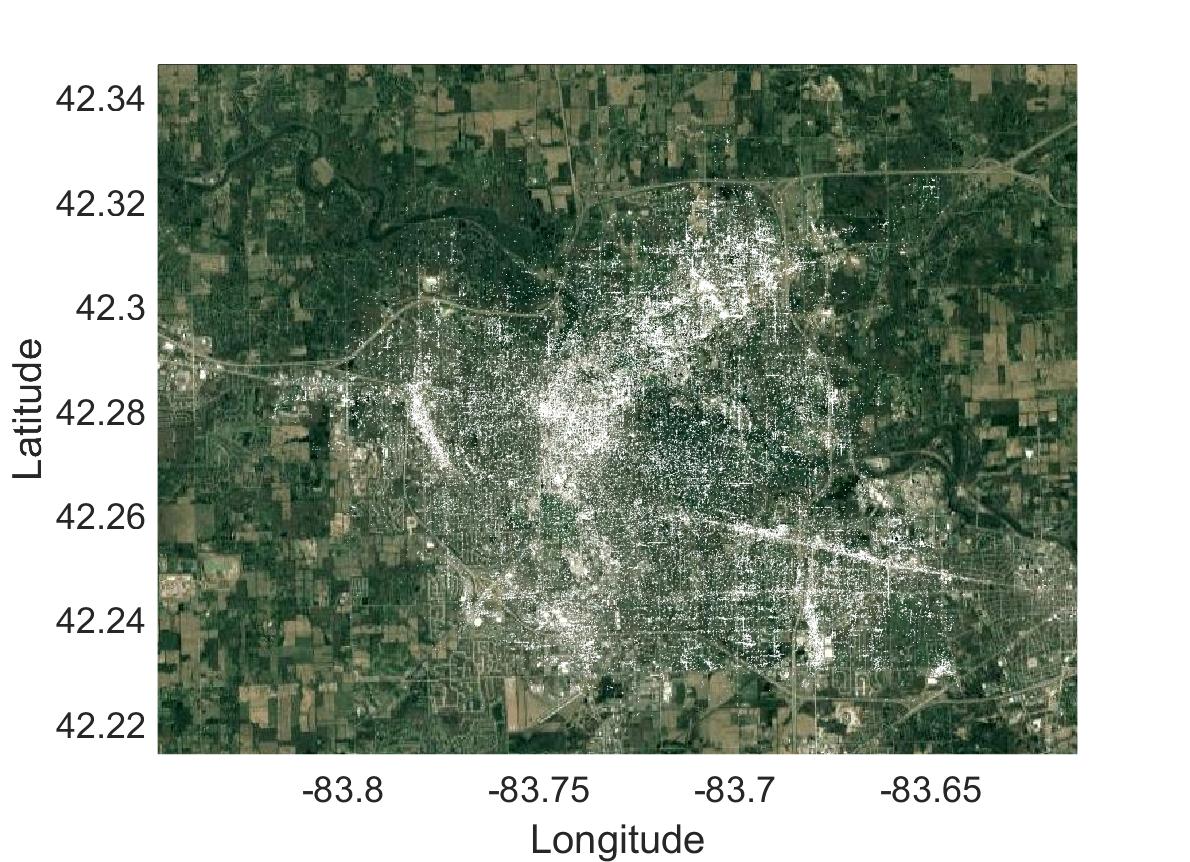}
    \caption{Distribution of the qualified trips in Ann Arbor. The white dots stands for the central points of the trips.}
    \label{fig:distribution}
\end{figure}
 
\subsection{Preprocess}

There are some noise and abnormal points, and therefore the raw data need to be processed before our mining process The categories of data error are as follows:
\begin{enumerate}
\item The position of vehicles in the abnormal area like Pacific. 
\item The discontinuous points in the data sequence.
\end{enumerate}

To eliminate the abnormal data, we sort all the sample point with a latitude limitation of [41.65, 44.5] and a longitude limitation of [-82.37, -86]. This is the range of the downtown area of Ann Arbor, Michigan.

The data discontinuity can be caused by sample lost, or that the preprocessing algorithm considers two different trips as one. This time discontinuity can cause velocity spike and also impede the following algorithm. Therefore we eliminate the whole trip even if there is only one point lost. Although this may cause the loss of the potential vehicle encounters data, it is tolerant as we have more than enough data.
After the filter, we get 20,000 qualified trips. The distribution of those trips are shown in Fig. \ref{fig:distribution}, in which the white dots stands for the central points of the trips. It is noted that those trips are not simultaneous while we present all of them in one image. As there is no discontinuity in the time series data, it is easy to calculate the duration and velocity of the trips.
\subsection{Temperal-Spatial Match}

\begin{figure}[b]
\centering
\subfloat[]{
\label{fig:timeslot} 
\includegraphics[width=1.5in]{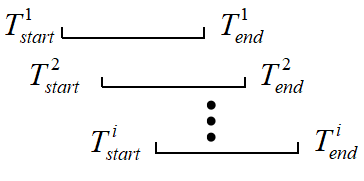}}
\subfloat[]{
\label{fig:queue} 
\includegraphics[width=1.5in]{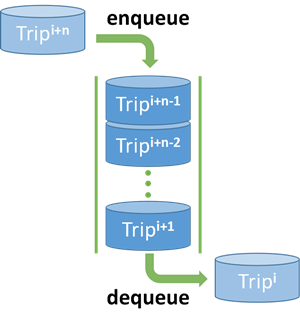}}
\caption{Queue-structure based matching. (a)The time slot reordered by Start Time and End Time. (b)Queue structure for the matching of the trip combination. }
\label{fig:spatialtempmatch} 
\setlength{\abovecaptionskip}{0pt}
\setlength{\belowcaptionskip}{0pt}
\end{figure}

The trip encounter is a combination of two trips sharing temporal and spatial intersection. We filter the scenes in which the trajectories have both temporal and spatial intersection. However, the combinatorial explosion is still a big problem that hinders the mining speed dramatically. For example, 300 trips can generate 44850 combinations, and this number will become 3.6 x 10$^{12}$ if we exploit the original scale of the database with 6 million trips in total, which is the total trip number in the Safety Pilot database. 

Here we overcome this problem by applying a queue-structure algorithm shown in Fig. \ref{fig:spatialtempmatch}. First, we reorder the trips by the start time and then the end time. Then we built a queue following the rules in Algorithm. \ref{alg:Match}. The running time decreases from previous 20 minutes to current 30 seconds. Most importantly, the complexity has been converted from a quadratic to a linear one, which is more suitable to be applied to a larger database. 
After the filter, 30,000 trip encounters are extracted from 300,000 trips.

 \begin{algorithm}[H]
 \caption{Match for the Trip Encounter}
 \label{alg:Match}
 \begin{algorithmic}[1]
 \renewcommand{\algorithmicrequire}{\textbf{Input:}}
 \renewcommand{\algorithmicensure}{\textbf{Output:}}
 \REQUIRE $TripSet$ 
 \ENSURE  $TripEncounter$
 \\ \textit{Initialisation} :
  \STATE $\{Queue\} \leftarrow Trip_{1}$ 
  \\ \textit{Match using queue} :
 \FOR{each trip $Trip_i$ in $TripSet$}
 \FOR{each trip $Queue_j$ in $\{Queue\}$}
 \IF{$Trip_i$ and $Queue_j$ have temporal and spatial intersection}
\STATE  $TripEncounter \leftarrow (Trip_{i}\ ID,Queue_{j}\ ID)$
\ELSE\STATE {Remove $Queue_{j}$ from $\{Queue\}$}
 \ENDIF
 \STATE $\{Queue\} \leftarrow Trip_{i}$
 \ENDFOR
 \ENDFOR
 \RETURN $TripEncounter$
 \end{algorithmic}
 \end{algorithm}
\subsection{Coarse filter}

Let the maximal velocity of two approaching vehicles be $V_{M1}$ and $V_{M2}$ respectively, as shown in Fig. 6. $t_{Intv}$ is the time duration when the vehicles first meet each other and then go back. If the distance of each sample time is bigger than $D$, there is no possibility that those vehicles encounters. Thus we can filter the vehicle encounter coarsely just by the general information. The algorithm of this part is shown in Algorithm \ref{alg:Coarse}. In this work, D and d are set to be 1000 meters and 100 meters respectively. After the query, 30\% events are eliminated in this process.

\begin{figure} [b]
    \centering
  \subfloat[]{%
       \includegraphics[width=0.8\linewidth]{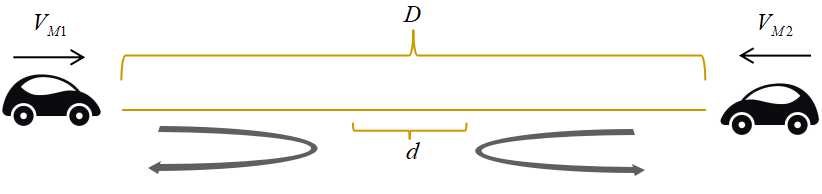}}\hfill
  \subfloat[]{%
        \includegraphics[width=0.8\linewidth]{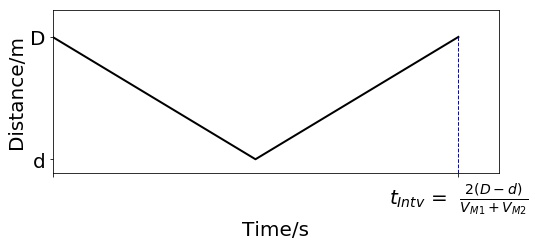}}\hfill
  
  \caption{(a)Scenario where two vehicles are approaching towards each other.(b)The distance of two vehicles. $t_{Intv}$ is used as the time interval for the coarse filter.}
  \label{fig:Approaching} 
\end{figure}

  \begin{algorithm}[H]
 \caption{Coarse Filter for Vehicle Encounter}
 \label{alg:Coarse}
 \begin{algorithmic}[1]
 \renewcommand{\algorithmicrequire}{\textbf{Input:}}
 \renewcommand{\algorithmicensure}{\textbf{Output:}}
 \REQUIRE $TripEncounterSet$ 
 \ENSURE  $VehicleEncounter$
 \FOR{each trip $TripEncounter_i$ in $TripEncounterSet$}
\STATE  Calculate the Time Interval: $t_{Intv}$ =  $\frac{2(D-d)}{V_{M1}+V_{M2}}$

 \STATE Re-sample the time series in coarse scale: $T_{Sample}=\{t_{Intv}, 2\times t_{Intv}, \dots, n\times t_{Intv}\}$
 \IF{min$\{Distance(t),t\in{T\_{Sample}}\}\geq{1000m}$}
\STATE  Eliminate $TripEncounter_i$

 \ENDIF

 \ENDFOR
 \RETURN $TripEncounter$ (filtered)
 \end{algorithmic}
 \end{algorithm}

\subsection{Fine filter}
In this last step, there are still 20,000 trip encounters remaining, and we calculate the distance of two vehicles at every sample point to find vehicle encounters. The total running time is 10 minutes on PC with Intel(R) Xeon(R) CPU E5-1620 v3.
The same method has been applied to a larger database, and 50000 events are queried in 10 hours.

\section{Clustering Method}
While the trajectories of single vehicle clustering are previously studied, there is no research focusing on the GPS data regarding the trajectories of two vehicles as a whole. For the former issue, the trajectories can be described as $T_1=\{(x_1,y_1), (x_2,y_2), \dots, (x_n,y_n)\}$ and $T_2=\{(u_1,v_1), (u_2,v_2), \dots, (u_n,v_n)\}$, which stands for the absolute coordinate of each vehicle. For the latter one, the trajectories can be described as $T_{12}=\{(x_1,y_1,u_1,v_1), (x_2,y_2,u_2,v_2), \dots, (x_n,y_n,u_n,v_n)\}$. What counts most is the relative position of two vehicles instead of the absolute position, and the absolute coordinate and direction are not the crux. To address that problem, we convert the trajectory to shape signature, by which the topology information is extracted. Then  K-means algorithm are used.

\subsection{Feature Extraction}
We use the relative heading angle and the relative distance as the feature. Fig. 7 illustrates the physical quantity we used. $\theta$ is the difference of the heading angle, the range of which is [0,180]. $L$ is the relative distance, the range of which is [0,100]. This method utilizes the primary and direct information which people concern. The relative heading angle suggests whether two vehicles are approaching, departing or at an intersection.

\begin{figure}
\centering
\includegraphics[width=0.8\linewidth]{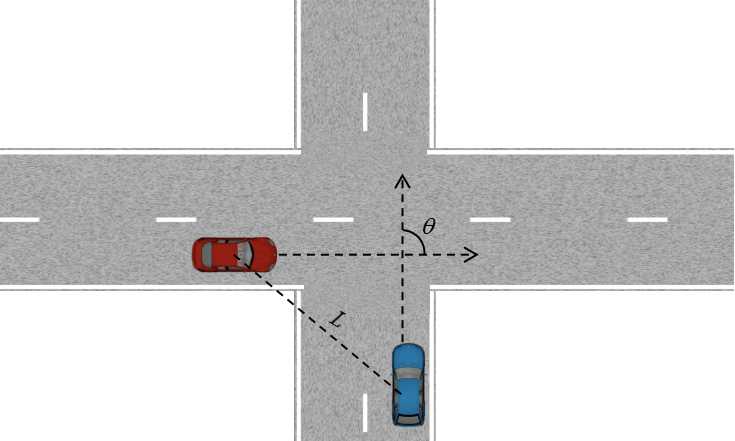}
\label{fig:DandA}
\caption{The relative distance and angle as features.}
\end{figure}


\subsection{Similarity Measurement}
The driving scenario may happen at the different local time. To cope with the local time shifting, two time-series are aligned for better similarity through the minimal number of delete. Here the Dynamic Time Warping (DTW) is a nonmetric similarity measurement that finds similar patterns between trajectories\cite{Magdy2016Review}. This method allows a sequence to stretch or shrink in order to get a better match with another one as shown in Fig. 8. The overall introduction about DTW can be found in \cite{Keogh2005Exact}.

\begin{figure}
\centering
\includegraphics[width=0.8\linewidth]{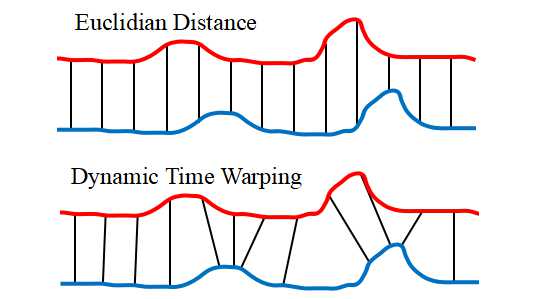}
\label{fig:DTW}
\caption{Note that while the two sequences have an overall similar shape, they are not aligned with the time axis. Euclidean distance will produce a pessimistic dissimilarity measure. Dynamic Time Warped alignment allows a more intuitive distance measure to be calculated.}
\end{figure}


\subsection{$k$-Means Clustering}
The $k$-means clustering is a very well-known unsupervised machine learning technique for classification. Given $n$ observations ($x_1$, $x_2$, $\dots$, $x_n$) and $k$ classes, the $k$-means clustering method classifies the observations into $k$ groups by solving a optimization problem
\begin{equation} 
\mathbf{c}^* = \argmin_\mathbf{c} \sum_{i = 1} ^{k}\sum_{j = 1}^{n}(\abs{x_{i} - v_{j}})^{2},
\end{equation}
where $\abs{x_{i} - v_{j}}$ is the Eucledian distance between the centroid and the observation \cite{kmeans}. This method tries to find centroids that can minimize the distances to all points belonging to its respective cluster. Thus the centroids are more representative among the surrounding data points in the same cluster.

\section{Results and Discussion}
Tested in a small dataset, the query time of the original mining method is 10 minutes, while our method just takes 30 seconds for this mining. Besides, we convert the quadratic computation to a linear. Therefore this method is also suitable for the large database without the problem occurred by the combinatorial explosion.

The primary driving feature in the real scene can be divided into three categories:
\begin{enumerate}
\item Car-following
\item Intersection
\item By-passing in opposite direction
\end{enumerate}

The feature chose can separate the primary driving scenario easily. Fig. 9 illustrates the feature space of car-following, intersection, and by-passing.

\begin{figure}[h]
\centering
\includegraphics[width=0.8\linewidth]{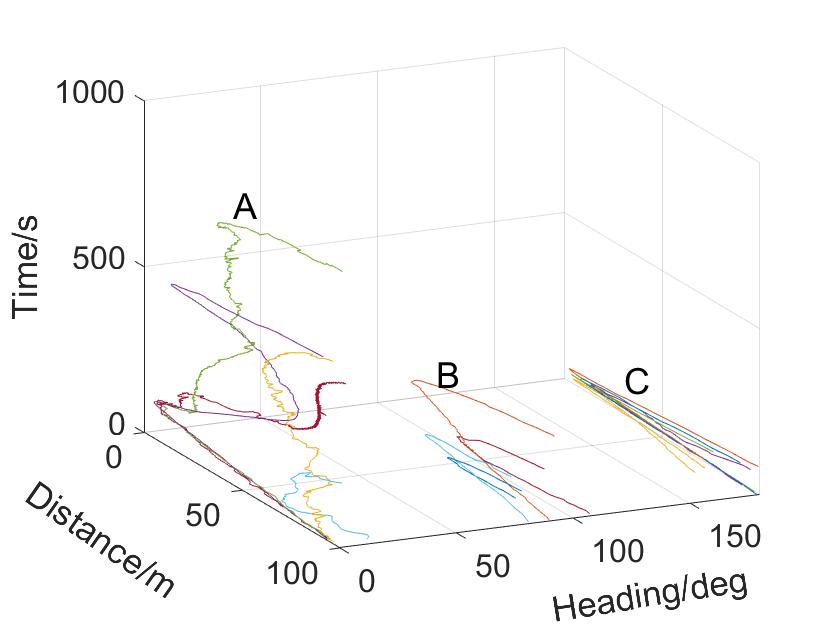}
\label{fig:FeatureSpace}
\caption{Three clusters in the feature space.}
\end{figure}

\begin{figure} [h]
    \centering
  \subfloat[]{%
       \includegraphics[width=0.5\linewidth]{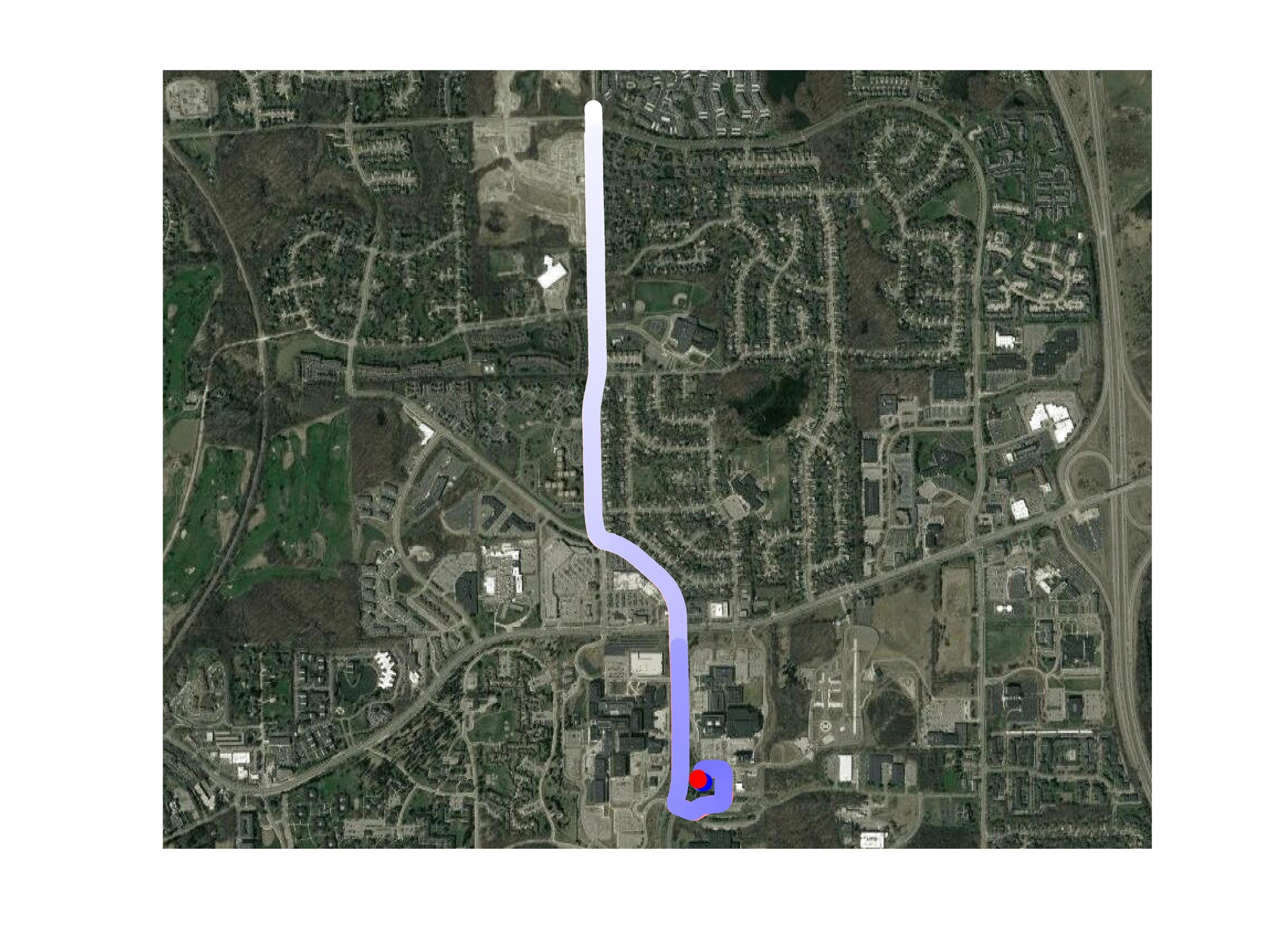}}\hfill
  \subfloat[]{%
        \includegraphics[width=0.5\linewidth]{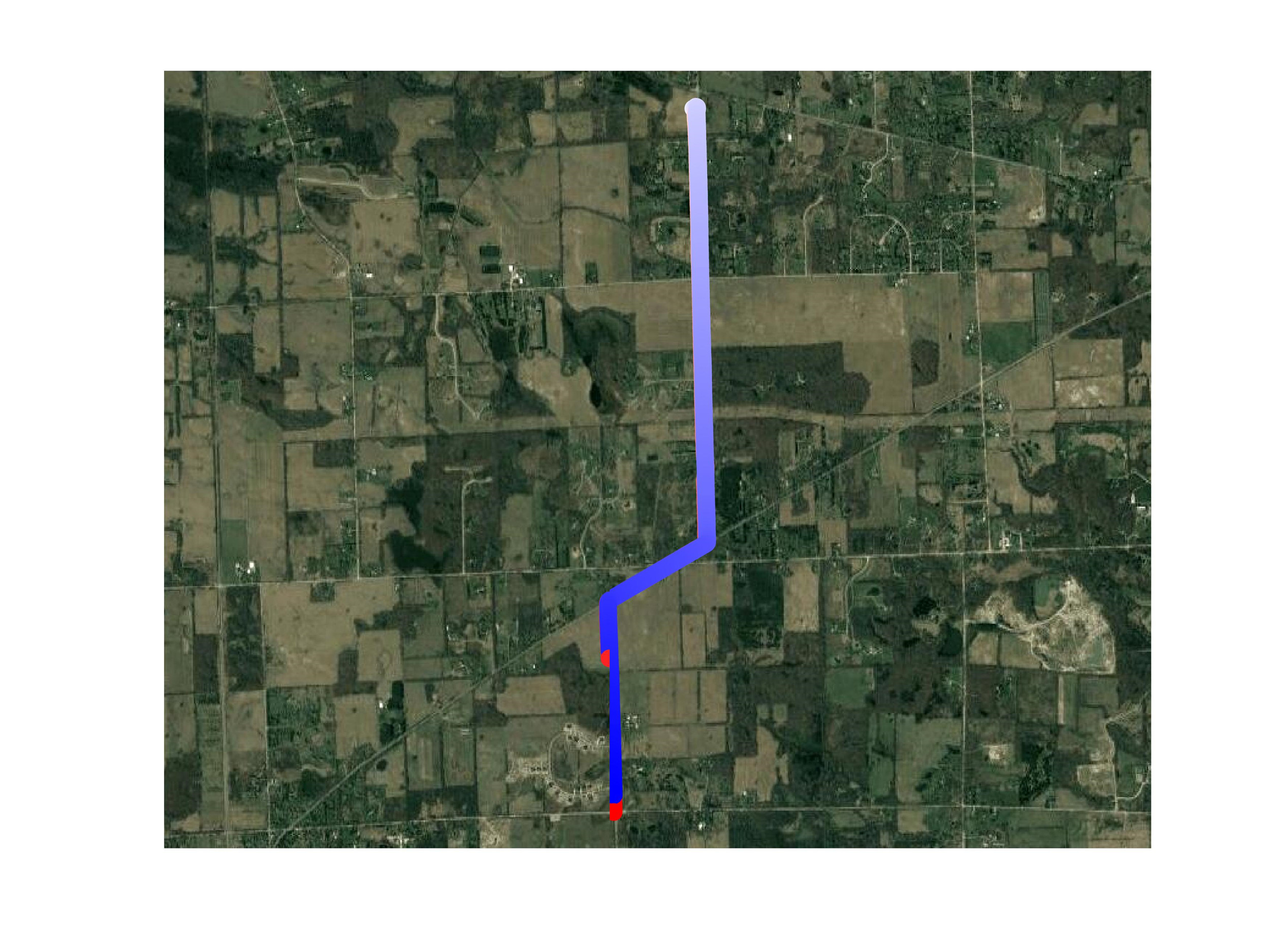}}\hfill
     \subfloat[]{%
       \includegraphics[width=0.5\linewidth]{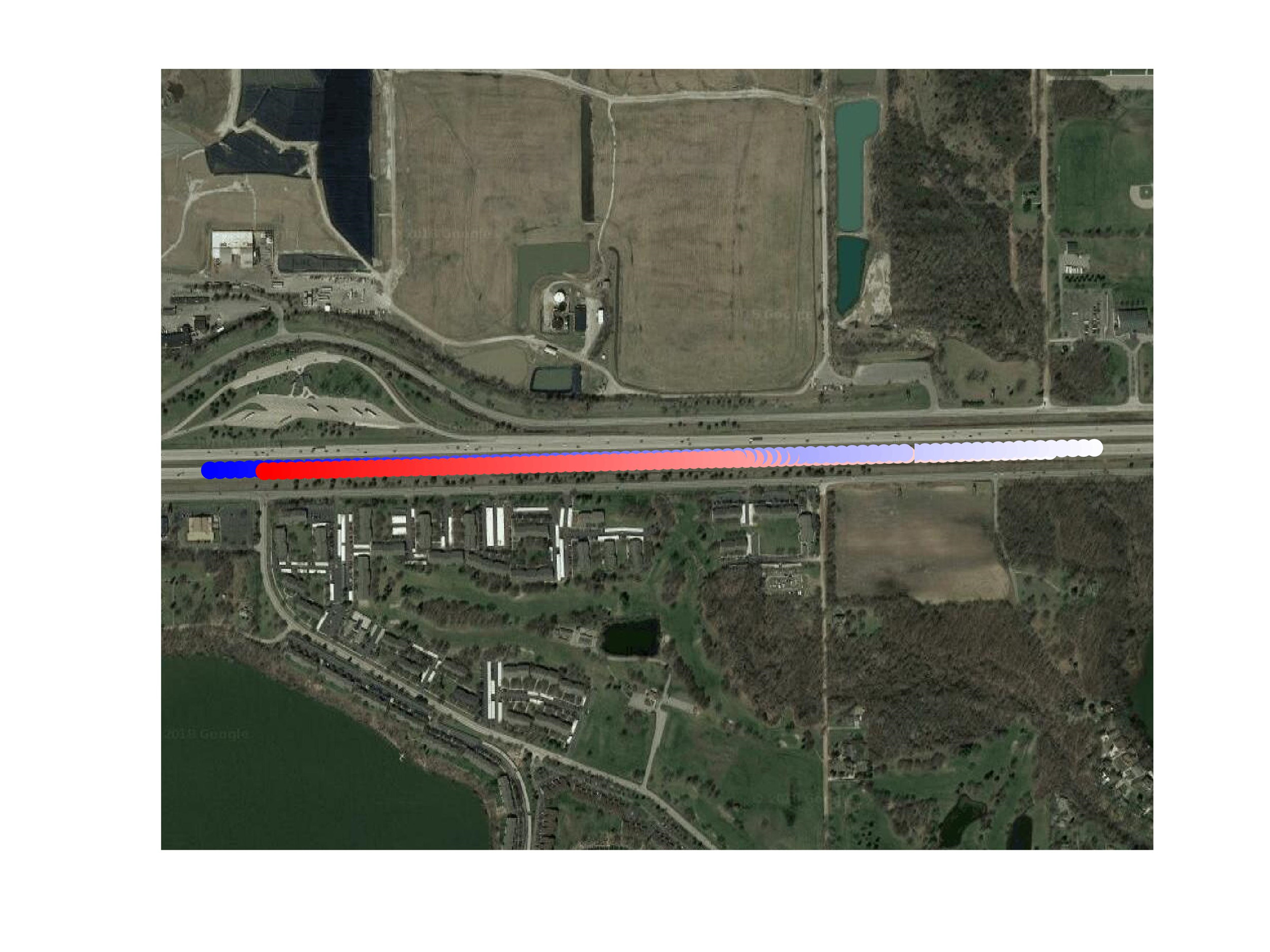}}\hfill
  \subfloat[]{%
        \includegraphics[width=0.5\linewidth]{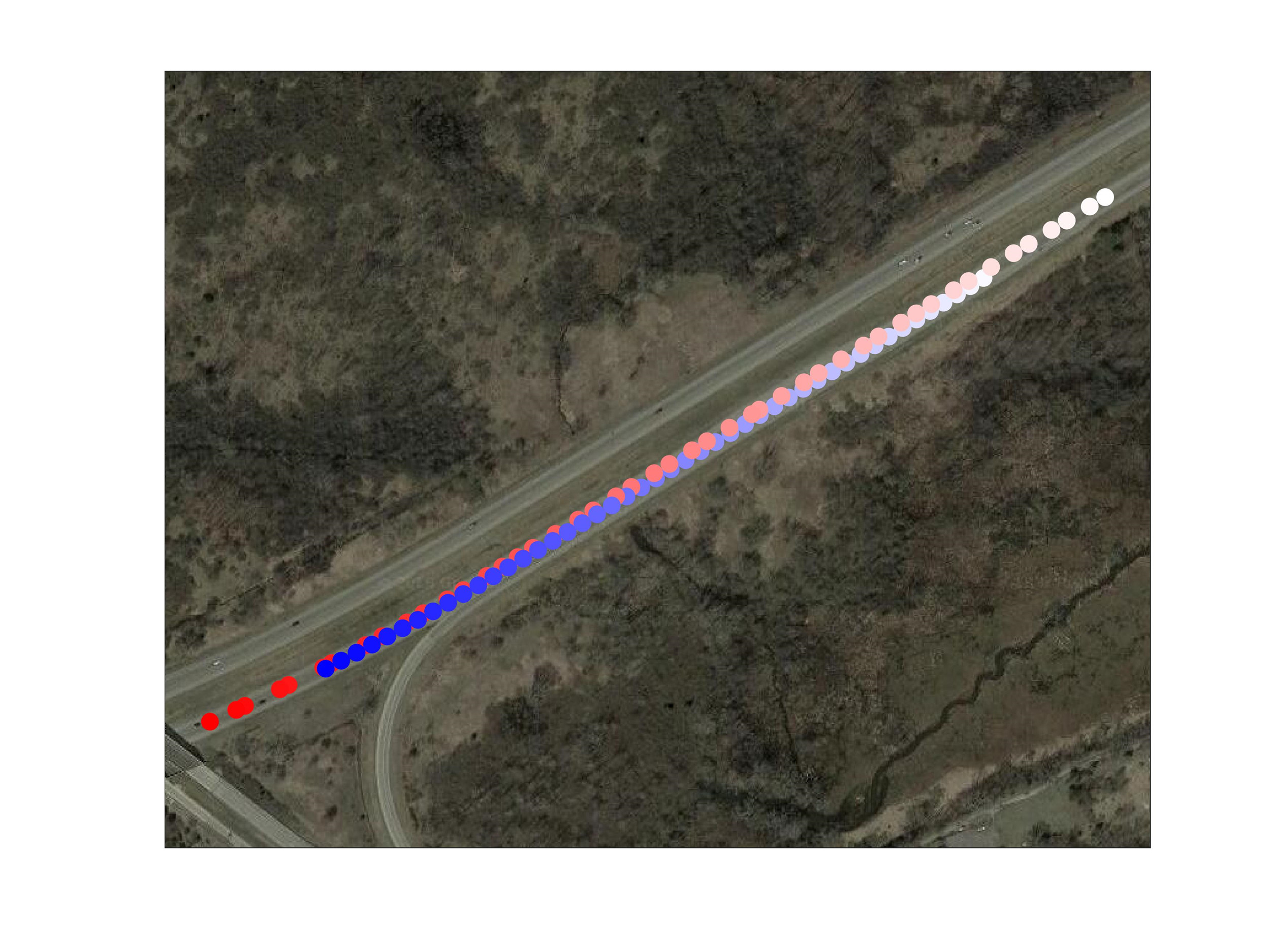}}
  \caption{Cluster A: one vehicle follows another.}
  \label{fig:CF} 
\end{figure}

Cluster A is the car-following scenario, as shown in Fig. \ref{fig:CF}. This part is distinct because normally this scenario takes obviously longer time than others, and the heading angle is approximately 0 degree. Although the distance pattern is exclusive for car-following, they can be aligned by the DWT only in some extent, therefore, is not a robust pattern.

\begin{figure} [h]
    \centering
  \subfloat[]{%
       \includegraphics[width=0.5\linewidth]{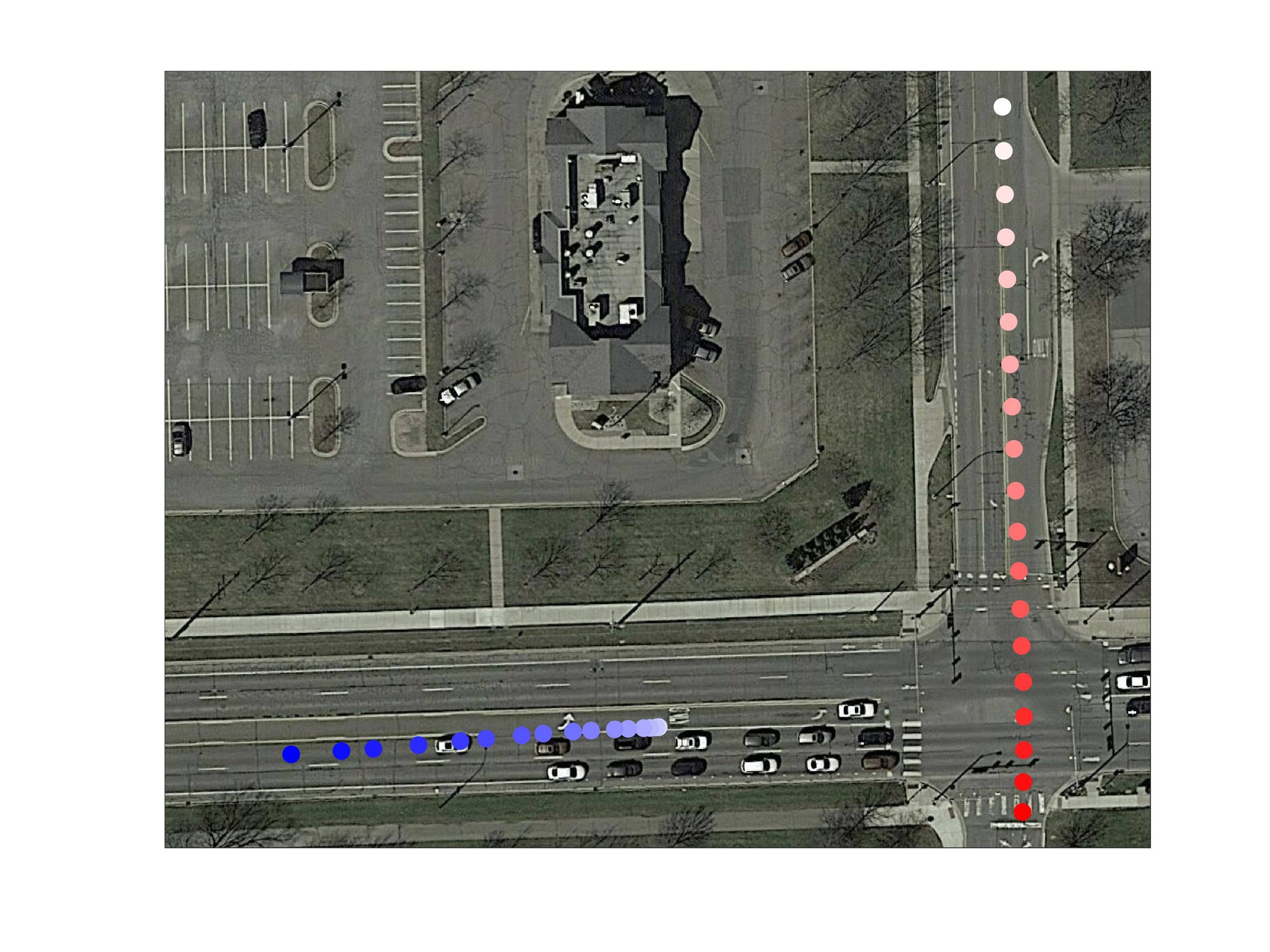}}\hfill
  \subfloat[]{%
        \includegraphics[width=0.5\linewidth]{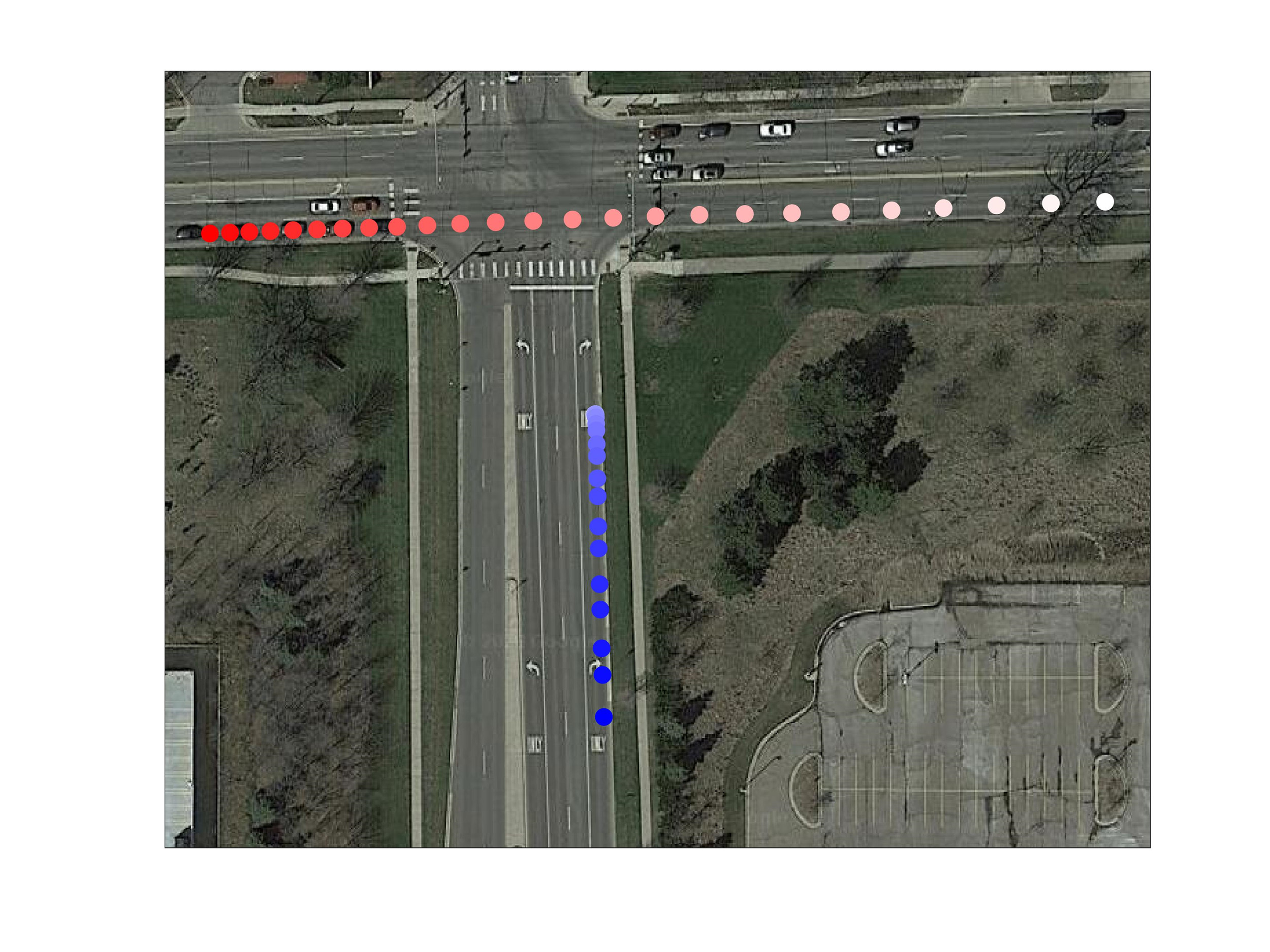}}\hfill
     \subfloat[]{%
       \includegraphics[width=0.5\linewidth]{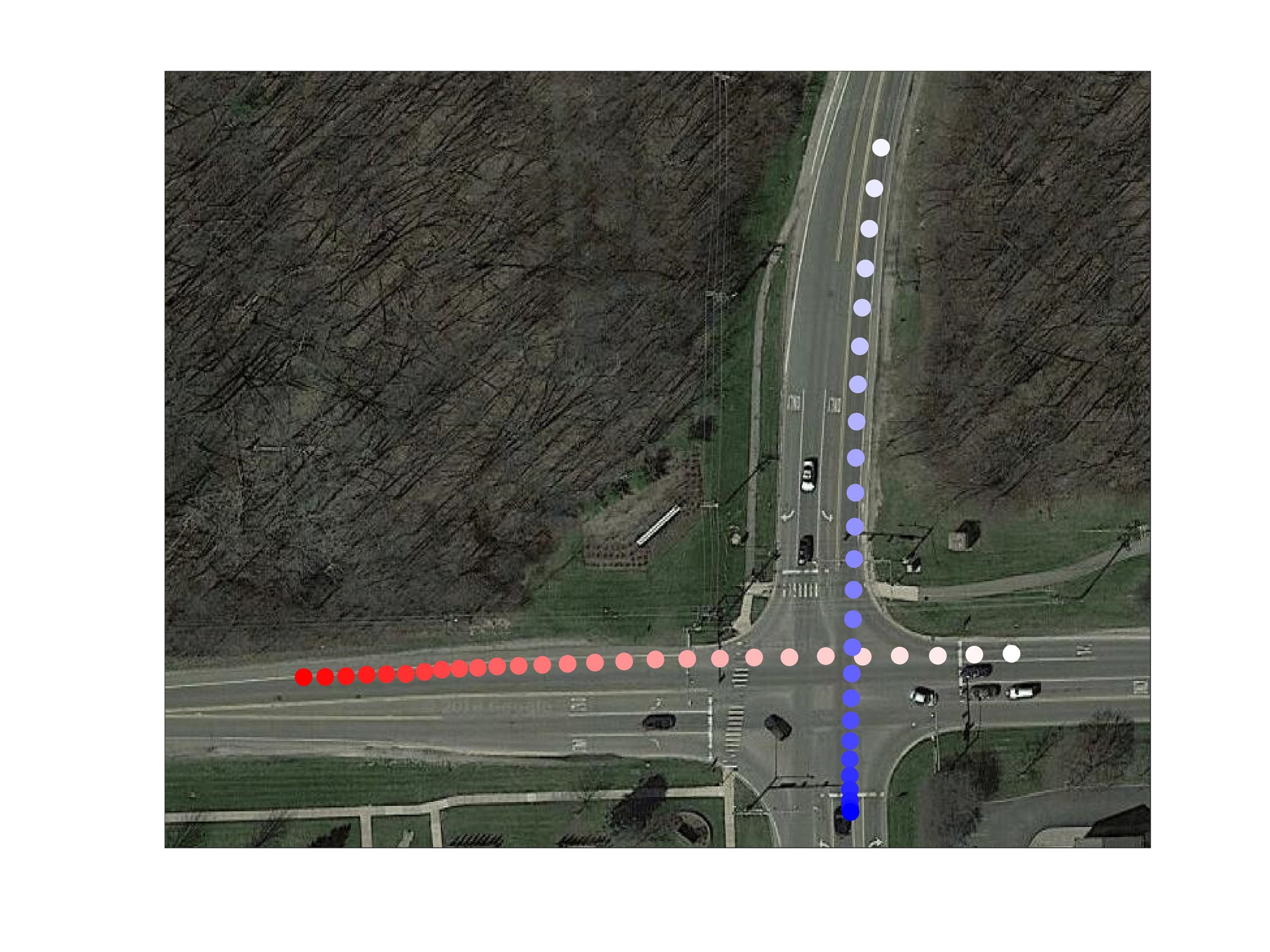}}\hfill
  \subfloat[]{%
        \includegraphics[width=0.5\linewidth]{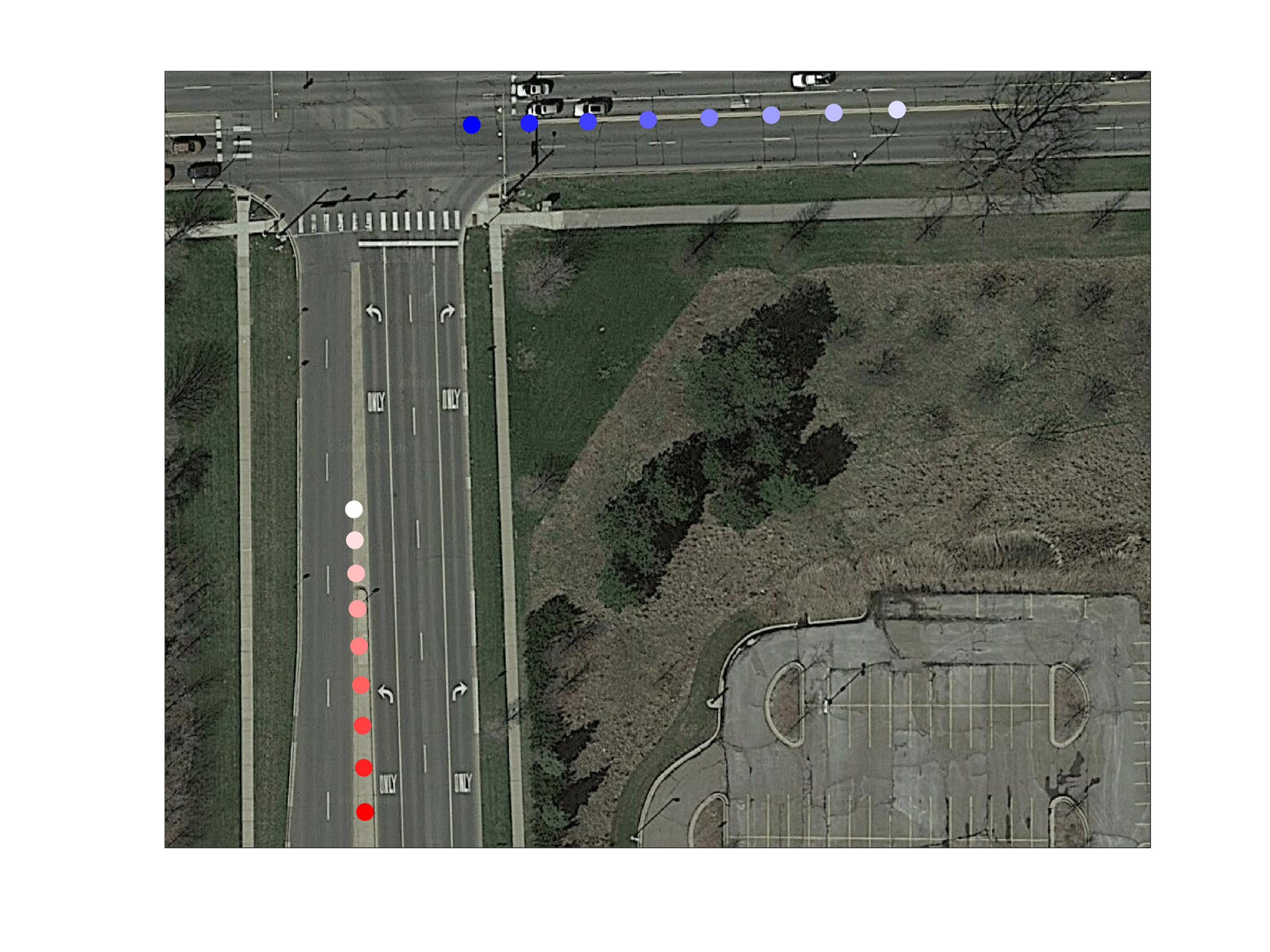}}
  \caption{Cluster B: one vehicle intersects with another in a road cross.}
  \label{fig:Intersect} 
\end{figure}

Cluster B is the intersection. The approximate 90 degrees heading angle difference is the primary characteristic, as shown in Fig. \ref{fig:Intersect}. Similarly, the trajectory is not aligned, and the DWT can adjust it very well.

\begin{figure} [h]
    \centering
  \subfloat[]{%
       \includegraphics[width=0.5\linewidth]{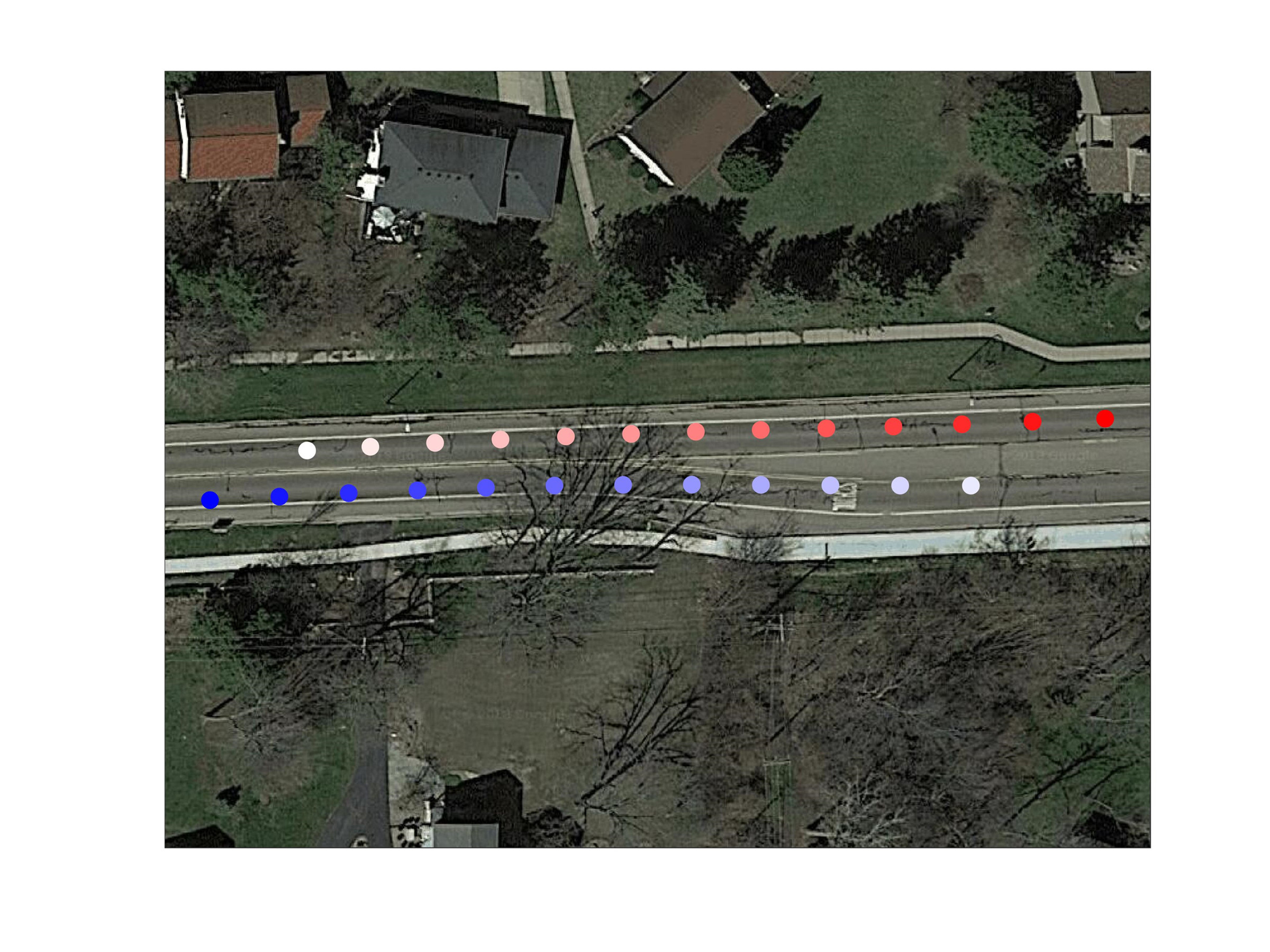}}\hfill
  \subfloat[]{%
        \includegraphics[width=0.5\linewidth]{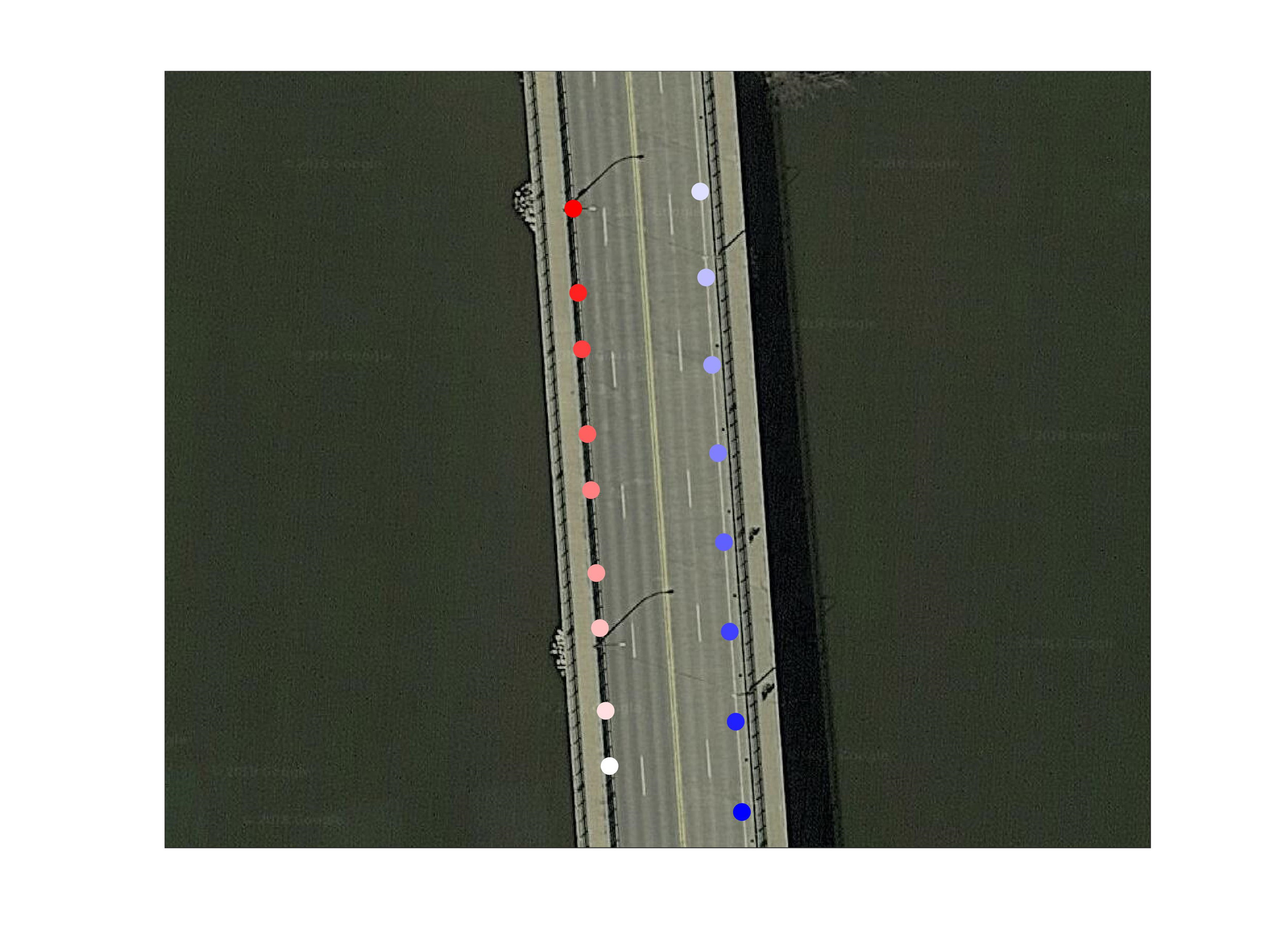}}\hfill
     \subfloat[]{%
       \includegraphics[width=0.5\linewidth]{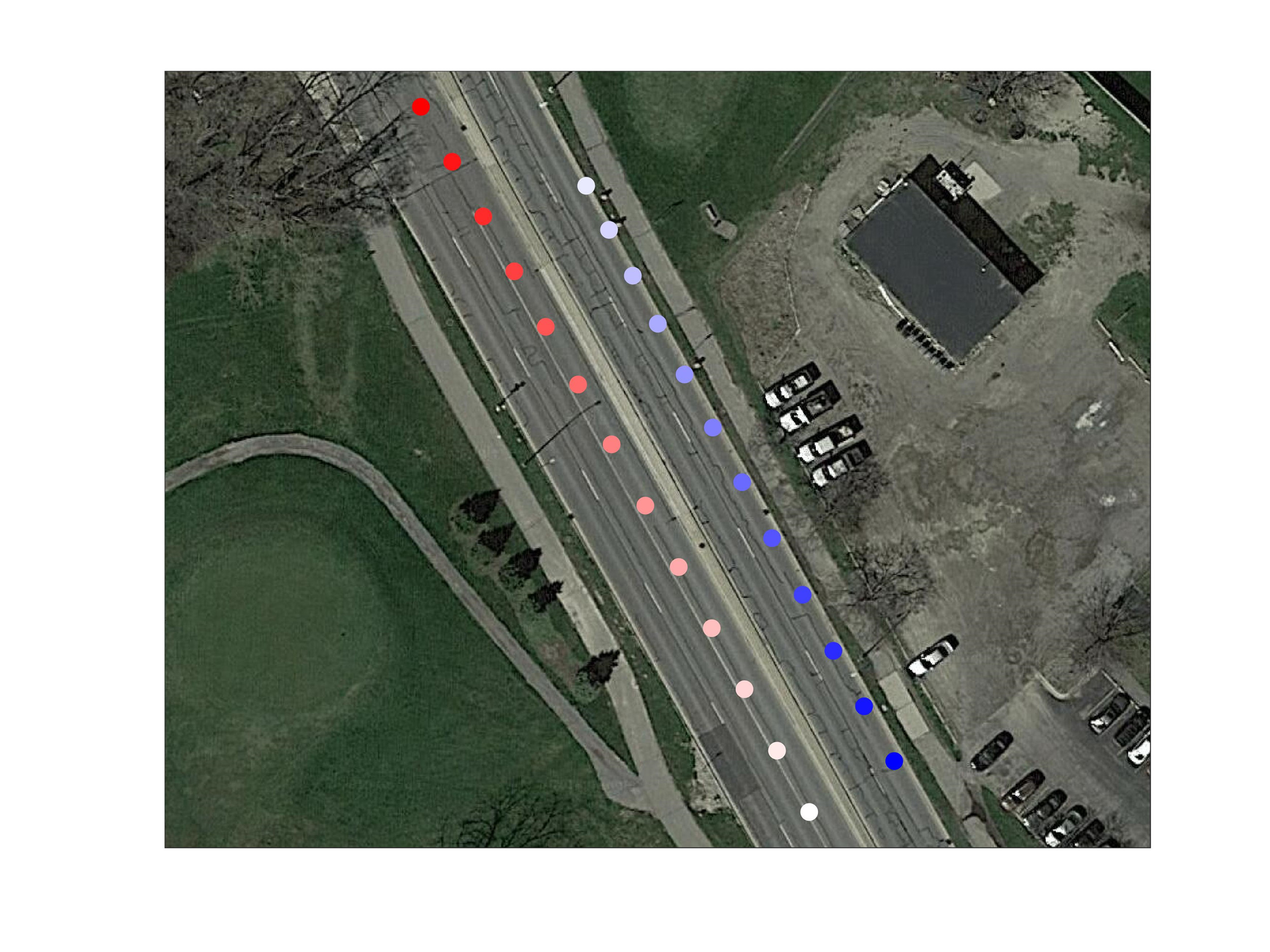}}\hfill
  \subfloat[]{%
        \includegraphics[width=0.5\linewidth]{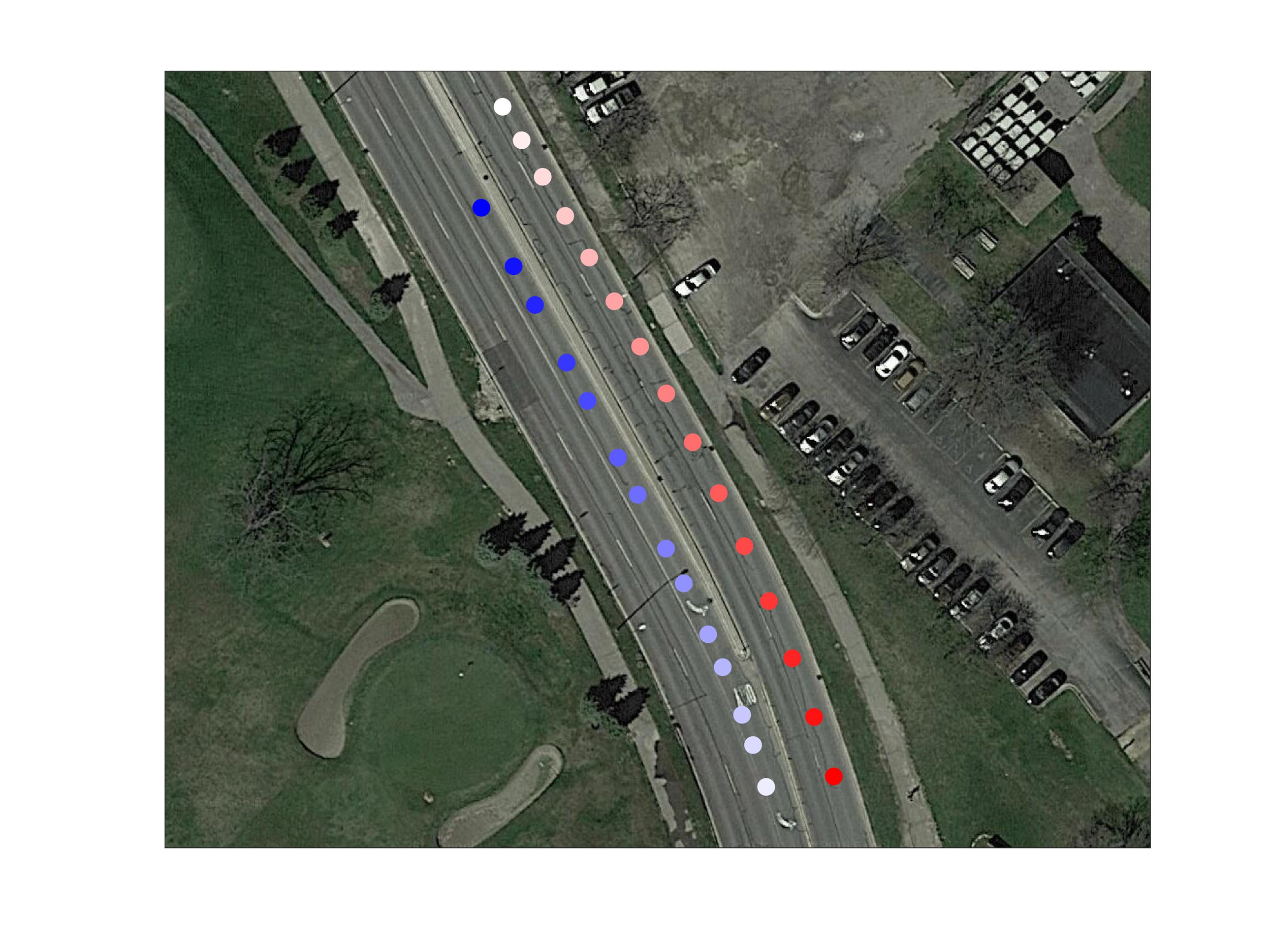}}
  \caption{Cluster C: one vehicle by-passes another.}
  \label{fig:BP} 
\end{figure}

Cluster C is the by-passing. The approximate 180 degrees heading angle difference is the primary characteristic, as shown in Fig. \ref{fig:BP}.

To evaluate the accuracy, the cluster results are label again manually, and then we calculate the number of events that are clustered correctly and wrongly. Thus the accuracy of each cluster is calculated as 
\begin{equation}
    \text{accuracy} = \frac{True}{True+False},
\end{equation}
where $True$ is the number of the events that are clustered correctly and $False$ is the number of the events that are clustered wrongly.

The method we proposed can cluster vehicle encounters that have similar geographic features but each cluster contains noisy data. Table~\ref{tab: accuracy} shows the accuracy of the clusters.

\begin{table}[h]
\centering
\caption{Clustering performance analysis in terms of accuracy.}
\label{tab: accuracy}
\begin{tabular}{c c c }
\hline
\hline
Cluster   & Accuracy \\
\hline
 Category A    &  56.7\%    \\
 Category B    &  72.6\%  \\
 Category C    &  79.4\%    \\
\hline\hline
\end{tabular}
\end{table}

The accuracy for car-following is low. It is because of the variant relative distance pattern. When the variance is small like those of cluster B and C, the DTW can align the local time shifting. However, DTW is not robust and is hard to handle the changeable time series. This problem can be overcome in the future by adding a penalty of the time duration, as the car-following have the longest encountering time. This is because the relative speed is low in car-following, and it can be used as a feature for clustering this scenario.

\section{Conclusion}
The paper provides a fast and general mining algorithm for mining vehicle encounter.  This is also the base for the scenario where three or more vehicles are included. The crux of this is the spatial-temporal filter and the coarse filter. The query contract show that it will decrease the query time dramatically and suitable for parallel computation and large database. Driving scenarios extracted from real-world data is also applicable to all kinds of Connected Vehicle evaluation. A more robust similarity measurement method is needed to achieve better clustering results.




\ifCLASSOPTIONcaptionsoff
  \newpage
\fi

\bibliographystyle{IEEEtran}
\bibliography{ref.bib}

\end{document}